\documentclass[journal,preprint]{vgtc_arxiv}

\ifpdf
  \pdfoutput=1\relax
  \usepackage{graphicx}
  \DeclareGraphicsExtensions{.pdf,.png,.jpg,.jpeg}
\else
  \usepackage{graphicx}
  \DeclareGraphicsExtensions{.eps}
\fi

\graphicspath{{figures/}{pictures/}{images/}{./}}

\usepackage{microtype}
\PassOptionsToPackage{warn}{textcomp}
\usepackage{textcomp}
\DeclareMathAlphabet{\mathcal}{OMS}{cmsy}{m}{n}
\usepackage{times}

\usepackage{cite}
\usepackage{tabu}
\usepackage{booktabs}
\usepackage{float}
\usepackage{longtable,booktabs}
\usepackage[ruled,vlined]{algorithm2e}
\usepackage{multirow}
\usepackage{amsmath}
\usepackage{amsfonts}
\usepackage{mathtools}
\usepackage[pagebackref=true,breaklinks=true, colorlinks,bookmarks=false,pagebackref,bookmarks]{hyperref}

\DeclareMathOperator*{\argmin}{argmin}

\newcommand{\norm}[1]{\left\lVert#1\right\rVert}
\renewcommand{\vec}[1]{\boldsymbol{#1}}

\newcommand{\x}{\vec{x}}

\newcommand{\I}{\vec{I}}

\def\eg{\emph{e.g}.} 
\def\ie{\emph{i.e}.}

\def\etal{\emph{et al}.}

\makeatletter
\newcommand{\thickhline}{
	\noalign {\ifnum 0=`}\fi \hrule height 1pt
	\futurelet \reserved@a \@xhline
}
\makeatother

\onlineid{0}

\preprinttext{}

\vgtccategory{Research}

\vgtcpapertype{application}

\title{LAPIG: Language Guided Projector Image Generation with Surface Adaptation and Stylization}
\supptitle{LAPIG: Language Guided Projector Image Generation with Surface Adaptation and Stylization}

\author{
  \authororcid{Yuchen Deng}{0009-0006-3893-955X},
  \authororcid{Haibin Ling}{0000-0003-4094-8413}, and
  \authororcid{Bingyao Huang}{0000-0002-8647-5730}
}

\authorfooter{
  \item
  	Yuchen Deng is with Southwest University.
  	E-mail: swudyc714@email.swu.edu.cn.
  \item
  	Haibin Ling is with Stony Brook University.
  	E-mail: hling@cs.stonybrook.edu.
  \item
    Bingyao Huang is with Southwest University.
  	E-mail: bhuang@swu.edu.cn. Corresponding author.
}

\abstract{
We propose LAPIG, a language guided projector image generation method with surface adaptation and stylization. LAPIG consists of a projector-camera system and a target textured projection surface. LAPIG takes the user text prompt as input and aims to transform the surface style using the projector.
LAPIG's key challenge is that due to the projector's physical brightness limitation and the surface texture, the viewer's perceived projection may suffer from color saturation and artifacts in both dark and bright regions, such that even with the state-of-the-art projector compensation techniques, the viewer may see clear surface texture-related artifacts. Therefore, how to generate a projector image that follows the user's instruction while also displaying minimum surface artifacts is an open problem. 
To address this issue, we propose projection surface adaptation (PSA) that can generate compensable surface stylization. We first train two networks to simulate the projector compensation and project-and-capture processes, this allows us to find a satisfactory projector image without real project-and-capture and utilize gradient descent for fast convergence. Then, we design content and saturation losses to guide the projector image generation, such that the generated image shows no clearly perceivable artifacts when projected. Finally, the generated image is projected for visually pleasing surface style morphing effects. The source code and video are available on the project page: \url{https://Yu-chen-Deng.github.io/LAPIG/}.
}


\teaser{
  \centering
  \includegraphics[width=\linewidth]{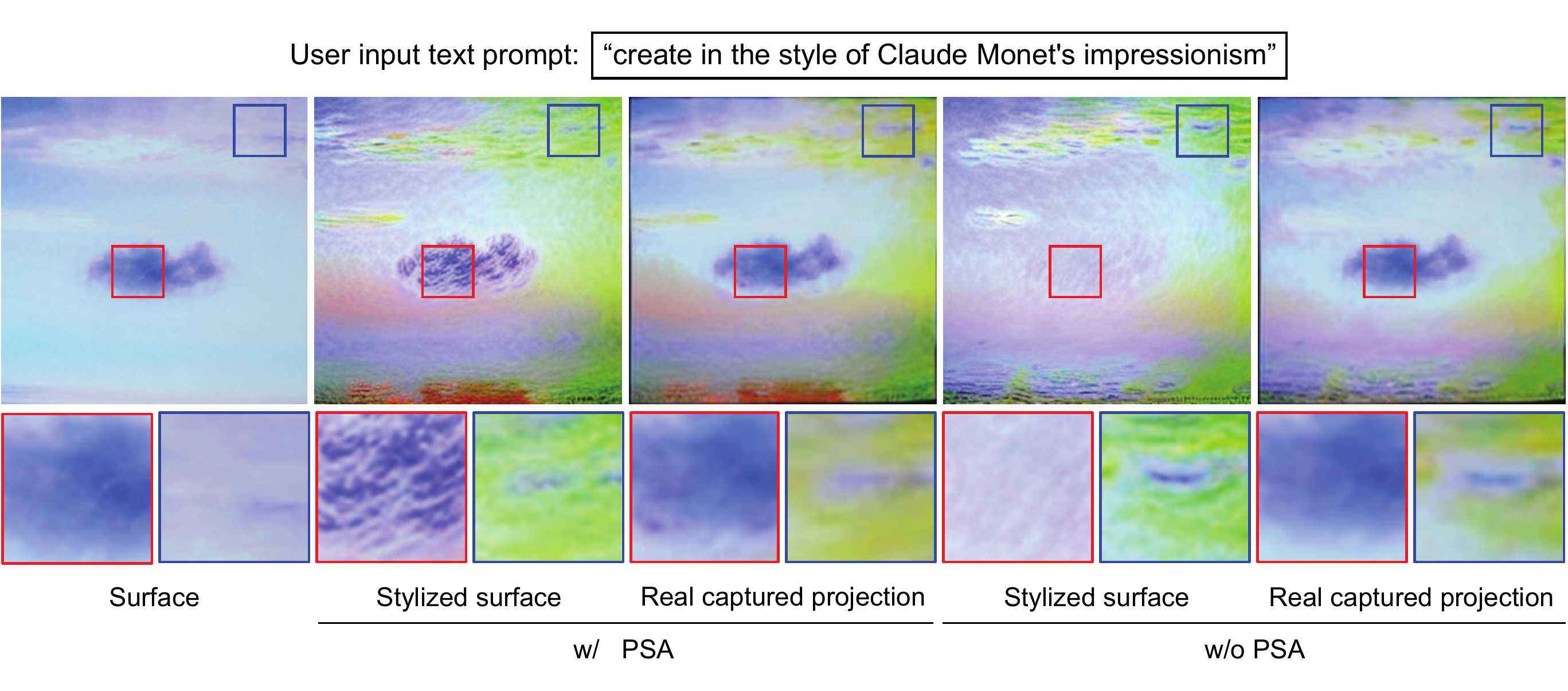}
    \caption{Language Guided Projector Image Generation with Surface Adaptation and Stylization (\textbf{LAPIG}): The user input text prompt to stylize the surface is shown on the top of the figure. The 1\textsuperscript{st} column shows the projection surface to be stylized. The 2\textsuperscript{nd} and 4\textsuperscript{th} columns show the stylized surface w/ or w/o our projection surface adaptation (\textbf{PSA}) approach. The 3\textsuperscript{rd} and 5\textsuperscript{th} columns present real captured scene under the compensated projection of the 2\textsuperscript{nd} and 4\textsuperscript{th} columns, respectively. 
    Clearly, our LAPIG can adapt the stylization to the surface by showing fewer discrepancies between the target stylized surface (the 2\textsuperscript{nd} column) and the real captured projection (the 3\textsuperscript{rd} column). 
    More results are provided in \autoref{sec:experiments} and \textbf{supplementary}.}
  \label{fig:teaser}
}

\begin{document}

\maketitle
\section{Introduction}
Projection mapping (PM)~\cite{iwai2006limpiddesk, Pjanic2018SeamlessMultiProjection, Kurth2018AutoCalibration, Asayama2018FabricatingDiminishable, Kurth2020RealTimeAdaptive, Akiyama2021RobustReflectanceEstimation, Hiratani2023ShadowlessProjectionMapping, Kageyama2024deblur, yasui2024pm} is a versatile technology that transforms the appearance of objects into dynamic displays by projecting digital images or videos onto them. This technique is utilized in various fields, including performing arts \cite{bandyopadhyay2001dynamicshaderlamps, bimber2005superimposing, Flagg2006painting, bermano2017makeuplamps, Kagami2019AnimatedStickies}, product design~\cite{bimber2005smartprojectors, iwai2006limpiddesk, Matsushita2011interactivebookshelf, CASCINI2020103308, erel2023neuralprojectionmapping, dong2023Calibration}, and interactive experiences~\cite{Marner2014projectorbasedar, Amano2019ManipulationOM, Miyatake2023HaptoMapping, iwai2006limpiddesk}, where it enhances the visual representation of objects with projected textures, colors, and other effects. By incorporating physical objects and environments into the display system, projection mapping increases user immersion and interaction, which is crucial for AR/VR applications.

Due to the complex environment, object material, and device optical properties, directly projecting desired patterns onto the target object may suffer from photometric and geometric distortions. Therefore, most projection mapping applications rely on projector compensation techniques \cite{Bimber2008calibration, grundhofer2018recent, grundhofer2015robust, huang2019compennet, huang2020CompenNeSt++, wang2023CompenHR, Kageyama2020ProDebNet, Li2024phycompen, Kageyama2024deblur} to cancel distortions. However, due to the projector's physical brightness limitation, not every projection surface is fully compensable, especially those extremely dark and bright regions, \textit{e.g.},  the dark cloud in the surface (the 1\textsuperscript{st} column) of \autoref{fig:teaser}. This is because the extremely dark/bright projection surface regions absorb/reflect more light, even when the projector brightness is set to the highest/lowest, the surface texture cannot be fully compensated.

To address this issue, we propose LAPIG, a language guided projector image generation method with surface adaptation and stylization. LAPIG consists of a language guided style transfer module (LGST) for user controllable surface stylization and a projection surface adaptation (PSA) module for compensable stylization. The key difficulty is how to combine the language guided style transfer and the compensable stylization. A naive approach is to repetitively generate different projector input images using LGST, then project and capture the generated images until a compensable stylization is achieved. However, this approach is time-consuming because the real project-and-capture process is involved. A smarter method would be to simulate the real project-and-capture and projector compensation processes using a mathematical model, \textit{e.g.}, a light transport matrix and its inverse, and then perform the similar random projector image generation as in the naive approach. However, this approach is still time-consuming due to exhaustive search and unstable convergence. 

We propose to address this issue using gradient descent rather than random search, for efficient convergence. To allow for gradient descent-based optimization, we first design two differentiable metrics for surface adaptation quality, \ie, projection consistency and color saturation losses. The first loss measures the similarity between the target stylization and the captured projection, and the second measures the projection and compensation color saturation losses in extremely dark and bright surface regions. Then, we pre-train two neural networks to simulate the projector compensation and project-and-capture processes, respectively. Finally, given an initial stylized image, we feed it to the compensation and project-and-capture modules to simulate the camera-captured compensation, which is used to calculate the projection consistency and color saturation losses. We iteratively backpropagate the loss gradient to update the LGST model parameters until the stylized surface is compensable.

The contributions of this work can be summarized in three aspects:

\begin{itemize}
\item The proposed LAPIG is the first language guided projector image generation method that can not only stylize the projection surface but also adapt to it.

\item LAPIG nontrivially integrates user language guided style transfer, project-and-capture simulation, and projector compensation, enabling gradient-based optimization for projector image generation.

\item We propose projection consistency and color saturation losses to measure compensation quality and guide PSA optimization, and they are expected to facilitate other projection mapping applications.
\end{itemize}

In the rest of the paper, we introduce the related work in \autoref{sec:related_work}, and describe the problem formulation and the proposed LAPIG in \autoref{sec:methods}. We show our system configurations and experimental evaluations in \autoref{sec:experiments}, discuss the paper in \autoref{sec:discussion}, and conclude the paper in \autoref{sec:conclusion}.

\section{Related Work}\label{sec:related_work}

\subsection{Projection mapping}

Traditional projection mapping (PM) techniques aim to alter the appearance of physical objects by projecting carefully designed patterns onto their surfaces. By doing so, these methods effectively modify the color, texture, and brightness of the object surfaces, serving as a widely adopted tool for spatial augmented reality (SAR). To achieve satisfactory visual effects in PM, projector compensation and projector-camera systems (ProCams) relighting (also referred to as project-and-capture simulation) are commonly employed.

\subsubsection{Projector compensation}

Projector compensation aims to counteract distortions caused by environmental factors, equipment, and projection surfaces, thus improving the viewer's perception of projection effects by producing a compensation image~\cite{bimber2005superimposing, bimber2005smartprojectors, Bimber2008calibration, grundhofer2015robust, grundhofer2018recent, huang2019compennet, huang2020CompenNeSt++, Kageyama2020ProDebNet, luo2021staypositive, wang2023CompenHR, Li2024phycompen}. Raskar \etal \cite{Raskar2003iLamps} achieved projection on non-flat objects using geometric corrections paired with camera radiometric calibration. In practice, while camera radiometric calibration is straightforward, recalibration becomes crucial when projector settings like brightness, contrast, or color profiles are altered, posing challenges for projection mapping applications that require frequent adjustment. To address these issues, Grundhöfer and Iwai \cite{grundhofer2015robust} introduced a robust photometric compensation method using a pixel-wise thin plate spline (TPS) to directly estimate the photometric compensation function from RGB sampling images, and waives radiometric calibration. Huang \etal \cite{huang2017radiometriccompensation} achieved visually satisfactory projections by exploring properties of the human visual system such as chromatic adaptation and perceptual anchoring. They also employed gamut scaling to mitigate clipping artifacts from camera and projector sensor constraints. Luo \etal~\cite{luo2021staypositive} also took advantage of the properties of the human visual system to generate a high-quality non-negative image, which may be applied to reduce projector compensation artifacts. Huang \etal \cite{huang2019compennet, huang2020CompenNeSt++} proposed a learning-based formulation of the project-and-capture process, and applied deep neural networks to learn photometric and geometric compensation functions. Wang \etal \cite{wang2023CompenHR} introduced CompenHR, a resource-efficient technique for high-resolution projector compensation. They also suggested a multi-threaded video compensation strategy to dynamically adjust the compensation parameters \cite{wang2024vicomp}. Kusuyama \etal~\cite{Kusuyama2024deblur} presented a projection system that optically eliminates shadows in PM, and Yasui \etal~\cite{yasui2024pm} proposed a novel approach using a mixed light field within the projector, enhancing PM efficiency in brightly lit settings. For more detailed reviews, see \cite{Bimber2008calibration, grundhofer2018recent, iwai2024projection}.

\subsubsection{ProCams relighting/Project-and-capture simulation}

ProCams relighting (or project-and-capture simulation) simulates the physical project-and-capture process of ProCams and allows adjustments to appearance editing based on inferred captured projection without actual project-and-capture. Early methods focus on light transport matrix (LTM)~\cite{Debevec2000ltm, Wang2009ltm, Chiba2018ltm}, which models the irradiance of each camera pixel as a linear combination of the radiances of all projector pixels. These methods can produce accurate global illumination, but are usually computationally intensive and require high initial conditions. To further improve relighting efficiency, Huang \etal \cite{huang2021deprocams} proposed an end-to-end trainable model that explicitly learns the photometric and geometric mappings involved in the project-and-capture process. Erel \etal \cite{erel2023neuralprojectionmapping} first formulated ProCams simulation using a neural radiance fields (NeRF) framework.

\begin{figure*}
\centering
\includegraphics[width=\linewidth]{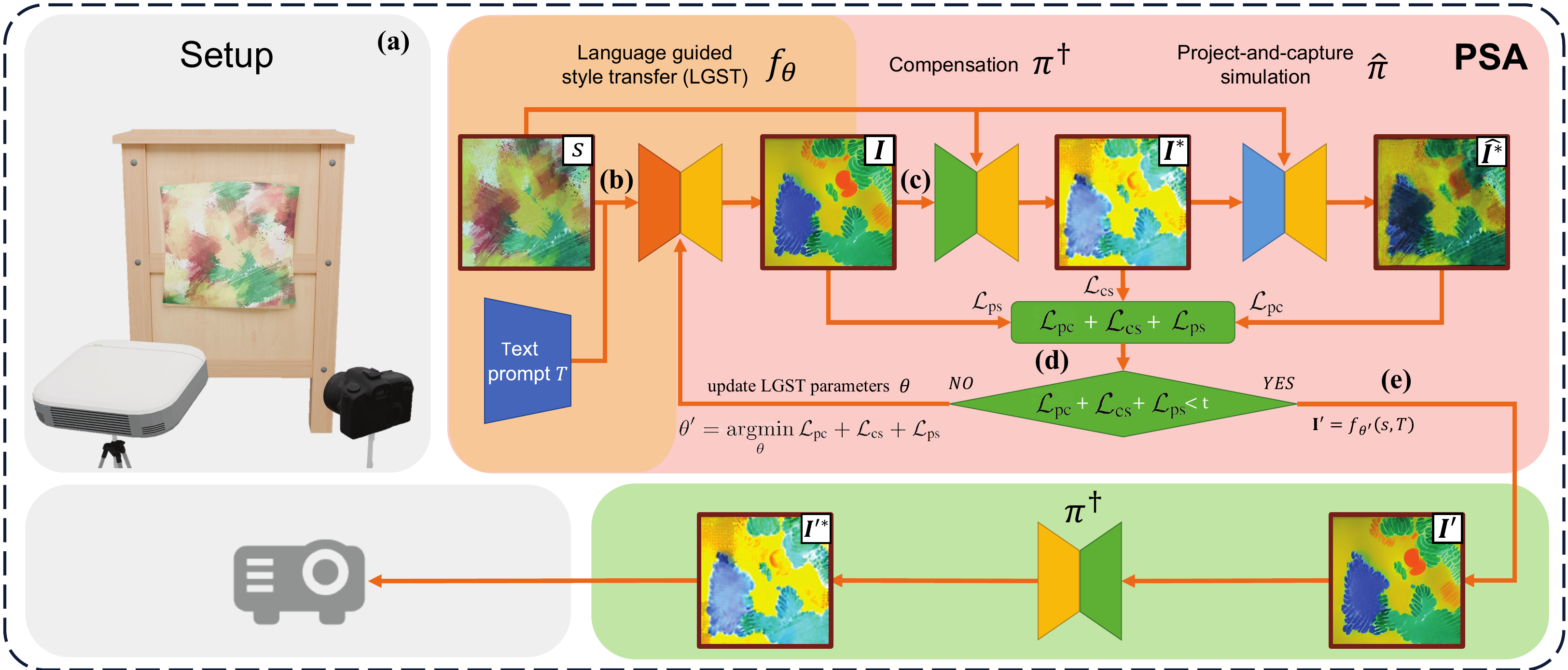}
\caption{System overview of our \textbf{LAPIG}. Note that we plot network in different colors to distinguish different modules utilized in PSA. (a) System setup consists of a projector, a camera, and a textured surface.  (b) First, the  surface $ s $ is captured by the camera, which is used in conjunction with text prompt $ T $ to generate a stylized surface image $ \I $. (c) Then, the surface  image $ s $ and its stylized image $ \I $ are input into the compensation network $\pi^{\dagger} $ to infer the compensated stylized image $ \I^* $. The compensated image is further input to the project-and-capture simulation network $\hat{\pi}$ to infer the camera-captured compensation effects $ \hat{\I^*} $.  The model architectures of $\hat{\pi}$ and $ \hat{\I^*} $ are shown in \autoref{fig:cmp-pc}. (d) Afterwards, the three compensation quality metrics/losses are calculated and backpropageted to update the LGST model parameters $\theta$. (e) Finally, the optimal stylized surface image $ \I' $ is generated using the updated LGST model, and after applying projector compensation, it is sent to the projector for real projection surface stylization.}
\label{fig:pipeline}
\end{figure*}

\subsubsection{Interactive projection mapping (IPM)}

Interactive projection mapping enables user feedback and control over the projected content. In scenarios such as art installations and museum displays, IPM not only enriches the presentation, but also provides the audience with the opportunity to participate in content creation. IPM can be categorized into three types. 

\vspace{1mm}\noindent\textbf{(1) Sensor-based interaction} captures user movements or expressions, allowing adjustment of projected content's appearance. Sensors~\cite{bermano2017makeuplamps, Peng2020high} and cameras~\cite{Flagg2006painting, iwai2006limpiddesk, Kagami2019AnimatedStickies, Asahina2021realistic, erel2024casperdpm} are used to capture the user's movements or expressions, allowing for the adjustment of the projected content's appearance. Amano~\cite{Amano2019ManipulationOM} introduced a light field projection method that enhances material perception based on the direction from which it is viewed. Recently, Miyatake \etal~\cite{Miyatake2023HaptoMapping} proposed a projection-based visuo-haptic AR system that allows independent rendering of visual and haptic content by embedding user-imperceptible tactile control signals in projected images. Erel \etal \cite{erel2024casperdpm} presented a novel approach to dynamically projecting 3D content onto the user's hands in real time.

\vspace{1mm}\noindent\textbf{(2) Touch-based interaction} provides users with direct control over projected content, allowing real-time adjustments via touch~\cite{Matsushita2011interactivebookshelf, Marner2014projectorbasedar, Punpongsanon2015softar}. Sato \etal~\cite{Sato2022ESP} proposed a high-speed projector-camera system that enables real-time interaction by creating optical illusions, dynamically altering the player's perception of a puck that is randomly hit during gameplay. 

\vspace{1mm}\noindent\textbf{(3) Text-based interaction} allows for dynamic and context-aware projection using verbal instructions or written descriptions. As diffusion models gain prominence, natural language also acts as a tool to alter and control projected content~\cite{erel2023neuralprojectionmapping}. 
To our best knowledge, there is no method that can perform language guided projector image generation with surface adaptation and stylization.

\subsection{Neural style transfer}

Our LAPIG leverages text prompt guided neural style transfer for surface adaptation and stylization, making it essential to review related work in this field.

\subsubsection{Non-text guided style transfer}

Neural style transfer (NST) was initially introduced by Gatys \etal~\cite{gatys2016styletransfer}, who proposed a method to blend content and style features extracted through a deep convolutional neural network (CNN). Building on this, Johnson \etal~\cite{johnson2016perceptuallossesrealtimestyle} introduced a model that performs fast style transfer using a pre-trained perceptual loss function, significantly speeding up the style transfer process. Other related works include style transfer models based on generative adversarial networks (GAN), such as CycleGAN~\cite{zhu2017cyclegan}, which can perform image-to-image translation without paired data. These models are entirely image-driven and do not incorporate any text prompts.

\subsubsection{Simple text guided style transfer}

As text-to-image generation models advanced, researchers began exploring the use of simple text prompts to guide the style and content of the generated images. An early example is by Reed \etal~\cite{pmlr-v48-reed16}, demonstrating the feasibility of generating image-matching natural language descriptions. Recently, transformer-based models~\cite{reddy2021dall, Avrahami2022SpaTextSR, Tao2023GALIPGA} have further expanded this concept, enabling the generation of diverse images based on simple text descriptions. These methods allow users to generate images based on natural language prompts, but often offer limited guidance on fine-grained style and detail.

\subsubsection{Fine-grained text guided style transfer}

To enhance the guidance of text in image generation and editing, researchers have developed models capable of fine-grained guidance over image style and content based on text prompts. CLIP \cite{radford2021learning}, a cross-modal contrastive learning model, significantly improved the alignment between text and images, laying the groundwork for subsequent text-driven image editing. Building on CLIP, models like StyleCLIP \cite{patashnik2021styleclip} have successfully combined CLIP with StyleGAN~\cite{Karras2019stylegan}, enabling detailed style editing of images according to precise text prompts. Further advancements include InstructPix2Pix~\cite{brooks2023instructpix2pix}, which extends the concept by providing an instruction-based interface for text-guided image editing.
InstructPix2Pix and Pix2Pix \cite{Isola2017pix2pix} both focus on image-to-image translation, with the former allowing users to apply fine-grained modifications to images according to textual instructions.
These models can not only generate images that match textual descriptions, but also adjust specific details such as facial expressions, colors, and textures based on user input.

\section{Methods}\label{sec:methods}

\begin{figure*}
\centering
\includegraphics[width=\linewidth]{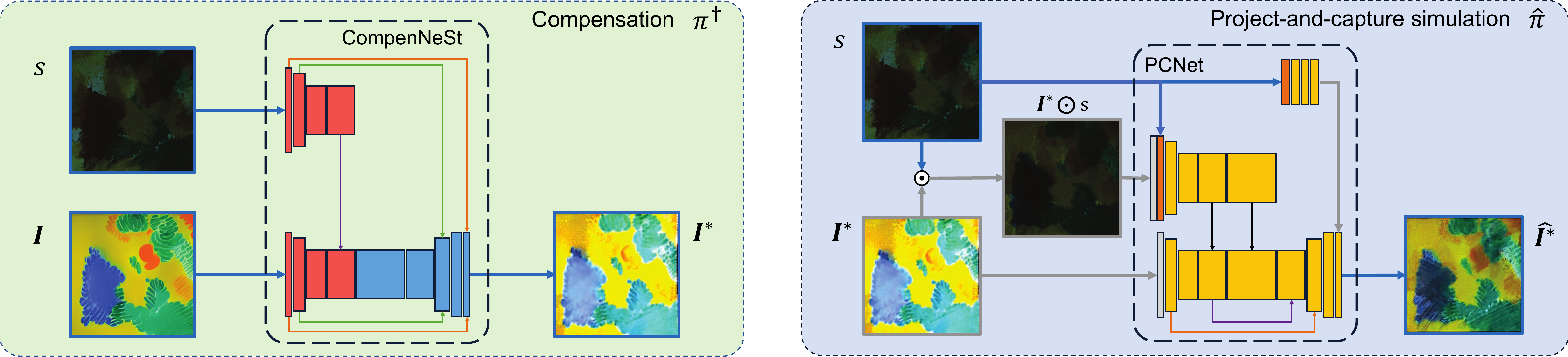}
\caption{Network architecture of projector compensation model $ \pi^{\dagger} $ and project-and-capture simulation model $ \hat{\pi} $.
CompenNeSt (identified by the light green block on the left) features a siamese encoder (red blocks with shared weights) and a decoder (blue blocks), performing the inverse mapping from the desired perceived image $\I$ to the projector compensation image $\I^*$. PCNet (represented by the light blue block on the right) utilizes a dual-branch encoder-decoder architecture to capture complex photometric transformations. Specifically, it calculates the rough shading $\I^* \odot s$ using  the camera-captured surface image $s$ and compensation image $\I^*$, and feed  them to the middle encoder branch. Likewise, $\I^*$ is passed to the backbone encoder branch. Skip connections between these two branches model photometric interactions across the three inputs at various levels. Additionally, $s$ is passed through three convolutional layers to the output layer. Ultimately, the backbone decoder fuses the feature maps into $\hat{\I^*}$, which simulates the camera-captured scene under the projection of $\I^*$.}
\label{fig:cmp-pc}
\end{figure*}

\subsection{Problem formulation}
We start by formulating the project-and-capture process of ProCams by expanding on the notation from~\cite{huang2021deprocams}. Denote the project and capture function as $\pi_\text{p}$ and $\pi_\text{c}$, respectively, and denote the projector input image as $\I$, the projection surface as $s$, then the camera-captured surface with superimposed projection $\tilde{\I}$ is given by:
\begin{equation}
    \tilde{\I} = \pi_\text{c}\left(\pi_\text{p}(\I), s\right)
\end{equation}
Denote the composite project-and-capture process and its inverse as $\pi$ and $\pi^{\dagger}$, respectively. Then, the above equation can be formulated as:
\begin{align}
    \tilde{\I} &= \pi\left(\I, s\right) \label{eq:pcp}\\
    \I  &= \pi^{\dagger}\left(\tilde{\I}, s\right)   \label{eq:cmp}
\end{align}
The inverse operation (\autoref{eq:cmp}) can be applied to projector compensation \cite{huang2020CompenNeSt++, huang2021deprocams} by replacing the input image $\tilde{\I}$ with a desired viewer perceived image. 

As shown in \autoref{fig:pipeline}, the goal of our LAPIG is to achieve projection surface stylization, according to the user input text prompt $T$:
\begin{equation}
    \I = f_\theta(s, T),  \label{eq:lgst}
\end{equation}
where  $\I$ is the desired viewer's perceived surface stylization effect (the 2\textsuperscript{nd} column of \autoref{fig:teaser}), and $f_\theta$ is a language guided neural style transfer model (LGST) parameterized by $\theta$, \eg, InstructPix2Pix \cite{brooks2023instructpix2pix}. Then, we apply projector compensation (\autoref{eq:cmp}) to $\I$:
\begin{equation}
    \I^* = \pi^{\dagger}\left(\I, s\right),
\end{equation}
and project the compensated image $\I^*$ to the projection surface, wishing the final camera-captured projection $\tilde{\I^*}$ matches the desired effect $\I$:
\begin{equation}
   \tilde{\I^*}= \pi\left(\I^* , s\right) \approx \I \label{eq:captured-compensation}
\end{equation}

\subsection{Projection surface adaptation (PSA)}

A straightforward solution for $\tilde{\I^*}$ in \autoref{eq:captured-compensation} relies on directly compensating the LGST ($f_\theta$) stylized image $\I$. Although LGST ($f_\theta$) can effectively edit or stylize an image based on text prompts, it lacks awareness of the actual ProCams configuration and may disregard the physical project-and-capture principles. As a result, the final captured stylized effect $ \tilde{\I^*}$ often displays noticeable surface artifacts and diverges significantly from the intended effect $\I$, particularly due to the complex environment, surface material, and brightness limitations of the projector, despite using advanced projector compensation methods (refer to the last two columns of \autoref{fig:teaser}). To address this issue, we propose conditioning the stylization of the projection surface on both the user's input text prompt and the quality of compensation (\autoref{fig:pipeline}). 

We first design three metrics/losses for compensation quality measure by:
\begin{align}
\mathcal{L}_\text{pc}:~  &\| \hat{\I^*} - \I\| \label{eq:pc}\\
\mathcal{L}_\text{cs}:~  &\| \max(\I^*-1,0)\|^2+\| \min(\I^*,0) \|^2 \label{eq:cs}\\
\mathcal{L}_\text{ps}:~  &\| \max(\I-\I_+,0)\|^2+\| \min(\I-\I_-,0) \|^2 \label{eq:ps} \\ 
\hat{\I^*} = &~\hat{\pi}(\I^*, s) = \hat{\pi}(\pi^{\dagger}(\I, s), s), \label{eq:simu-cap-cmp}
\end{align}
where $\hat{\I^*}$ is the simulated camera-captured projection of $\I^*$, and $\hat{\pi}$ is a model that simulates the real project-and-capture process $\pi$ (\autoref{fig:cmp-pc} right block).  $\I_+$ and $\I_-$ are two projector surface images captured under the brightest and darkest projector illumination to measure the projector's physical brightness range. $\mathcal{L}_\text{pc}$, $\mathcal{L}_\text{cs}$ and $\mathcal{L}_\text{ps}$ are projection consistency, compensation saturation, and projection saturation, respectively. In particular, $\mathcal{L}_\text{pc}$ measures the similarity between the simulated and desired surface stylization effects. $\mathcal{L}_\text{cs}$ measures the color saturation error due to projector compensation, \ie, pixel values outside the plausible RGB value range [0, 1] are counted as saturation errors. Similarly, $\mathcal{L}_\text{ps}$ measures the color saturation error due to the projection, \ie, any pixel value outside the range of the darkest ($\I_-$) and brightest ($\I_+$) images captured by the camera is unattainable due to the projector's brightness limitations, and should also be regarded as saturation errors.

We then propose a gradient-based iterative refinement method to optimize the language guided neural style transfer model parameters $\theta$.  This ensures that the stylized surface $\I'$ aligns with the input language guidance and remains compensable:
\begin{align}
   &\theta'= \argmin_{\theta}(\mathcal{L}_\text{pc}+\mathcal{L}_\text{cs}+\mathcal{L}_\text{ps})\label{eq:loss-minimization}\\
   &\I' = f_{\theta'}(s,T)\\
   &\I'^* = \pi^{\dagger}(\I')
\end{align}
Finally, the compensated surface stylization image $\I'^*$ is projected for surface stylization. The termination criterion for the gradient descent-based optimization  in \autoref{eq:loss-minimization} is determined by a loss threshold $t$. The detailed PSA algorithm is shown in \autoref{alg:pas}.

\setlength{\algomargin}{0em}

\begin{algorithm}[t]

\SetAlCapHSkip{0em}
\SetAlgoLined
\SetKwComment{Comment}{$\triangleright$\ }{}
\SetSideCommentLeft
\SetAlgoLongEnd
\DontPrintSemicolon
\SetKwInput{Input}{Input}
\SetKwInOut{Output}{Output}
\SetKw{KwOr}{or}
\SetKwRepeat{Do}{do}{while}

\caption{PSA: Projection Surface Adaptation.}\label{alg:pas}

\Input{\;
$s$: camera-captured projection surface under gray illumination\;
$T$: user input text prompt\;
$t$: PSA loss threshold for optimization termination\;
$\beta$: step size in minimizing the PSA losses\;
}

\Output{
$\I'^*$: compensated surface stylization image}

\vspace{1.5mm}Initialize $\theta' \gets \theta$\;

\Do{$\mathcal{L}_\text{pc}+\mathcal{L}_\text{cs}+\mathcal{L}_\text{ps} > t$}{
	$\I \gets f_{\theta'}(s, T)$ \Comment*[r]{surface stylization}\;
    $ \I^* \gets \pi^{\dagger}\left(\I, s\right)$ \Comment*[r]{compensation} \;
    $\hat{\I^*} \gets \hat{\pi}(\I^*, s)$ \Comment*[r]{simulate project-and-capture}  \;
$\mathcal{L}_\text{pc} \gets  \| \hat{\I^*} - \I\|$ \;
$\mathcal{L}_\text{cs} \gets  \| \max(\I^*-1,0)\|^2+\| \min(\I^*,0) \|^2$ \;
$\mathcal{L}_\text{ps} \gets  \| \max(\I-\I_+,0)\|^2+\| \min(\I-\I_-,0) \|^2$ \;
    $g\gets\nabla_{\theta'} (\mathcal{L}_\text{pc}+\mathcal{L}_\text{cs}+\mathcal{L}_\text{ps})$ \Comment*[r]{minimize PSA loss}\;
    $\theta'\gets \theta'+\beta*\frac{g}{\norm{g}_2}$\;
  }

\KwRet $\I'^*\gets \pi^{\dagger} (f_{\theta'}(s, T))$\;
\end{algorithm}

\subsection{Project-and-capture simulation and compensation}

The designs of the project-and-capture and the projector compensation models in \autoref{eq:pcp} and \autoref{eq:cmp} are crucial to our LAPIG, as their simulation accuracy of the real project-and-capture and compensation processes directly affects the final surface adaptation quality and viewer's perceived effects.

Inspired by the previous study \cite{huang2022spaa} that the project-and-capture process $\pi$ can be modeled by a neural network named PCNet ($\hat{\pi}$), we modify PCNet \cite{huang2022spaa} by removing the direct light mask requirement and applying it to our project-and-capture simulation, as shown in the right block of \autoref{fig:cmp-pc}. In particular, 
PCNet utilizes a dual-branch encoder-decoder architecture to capture complex photometric transformations, allowing it to model the photometric changes introduced by the projection. By this characteristic, PCNet can accurately simulate the real-world project-and-capture process.

For the projector compensation network $\pi^{\dagger}$, we modified the photometric compensation subnet (CompenNeSt) of the full projector compensation model CompenNeSt++ \cite{huang2020CompenNeSt++}. Specifically, CompenNeSt can correct distortions in color and brightness. As depicted in the left section of \autoref{fig:cmp-pc}, it utilizes a siamese encoder-decoder setup with two branches addressing the surface and projected images. The encoder (red blocks) in each branch shares weights and extracts features from the surface and projected images, merging them through skip connections. This alignment and combination enable the decoder (blue blocks) to reconstruct an image that compensates for photometric distortions such as color intensity, contrast, and brightness variances.

The effectiveness of the projector compensation network $\pi^{\dagger}$ and the project-and-capture simulation model $\hat{\pi}$ is shown in \autoref{sec:exp_simlulated_cmp} and \autoref{sec:exp_pcnet_losses}.

\subsection{Training details}

We implemented PSA using PyTorch \cite{paszke2017automatic} and optimized it using Adam optimizer \cite{kingma2015adam}. For LGST, we used the pre-trained weights of InstructPix2Pix \cite{brooks2023instructpix2pix} and optimized its parameters with an initial learning rate of $\beta=0.001$, and it decayed by a factor of 5 for every 50 iterations. For the compensation network \ie, CompenNeSt $\pi^{\dagger}$ in PSA, the hyperparameters for the training model are: learning rate of 0.001, batch size of 8, and 800 iterations, which took about 3.5 minutes to finish training on an Nvidia GeForce RTX 4060 laptop GPU. For the project-and-capture simulation network \ie, PCNet $\hat{\pi}$, the training hyperparameters are as follows: the learning rate is 0.001 with 1,500 training iterations. For training sizes of 8 and 48, the batch size is set to 8, and for a training size of 125, it is 24. Training on an Nvidia GeForce RTX 3090 GPU was completed in approximately 1 minute.

\section{Experiments}\label{sec:experiments}

\subsection{System configuration}
Our ProCams consists of a Canon 600D camera and an EPSON EB-C2050WN projector and their resolutions are $1280 \times 720$ and $640 \times 480$, respectively. Considering projector-camera synchronization, we manually set a 130-ms delay between projection and the capture operation, and in total it takes about 200 ms to capture a frame.  
We deploy LAPIG on a laptop with Intel Core i7-13650HX CPU, 16GB memory, and an Nvidia GeForce RTX 4060 Laptop GPU for both surface stylization and ProCams control.

\subsection{Datasets}
Our PSA involves both projector compensation $\pi^{\dagger}$ and project-and-capture simulation $\hat{\pi}$. Although there exist some datasets for projector compensation or project-and-capture simulation \cite{wang2023CompenHR, huang2020CompenNeSt++, Li2024phycompen, huang2021deprocams}, to the best of our knowledge, there is no public dataset to evaluate both methods simultaneously. Therefore, we captured a real dataset with 10 setups with real compensation images generated by \cite{huang2020CompenNeSt++}. In each setup, at least one of the texture, material, and surface geometry is different. Each setup has 180 image pairs for training and 20 for testing. 

\begin{figure}[!htbp]
\centering
\includegraphics[width=\linewidth]{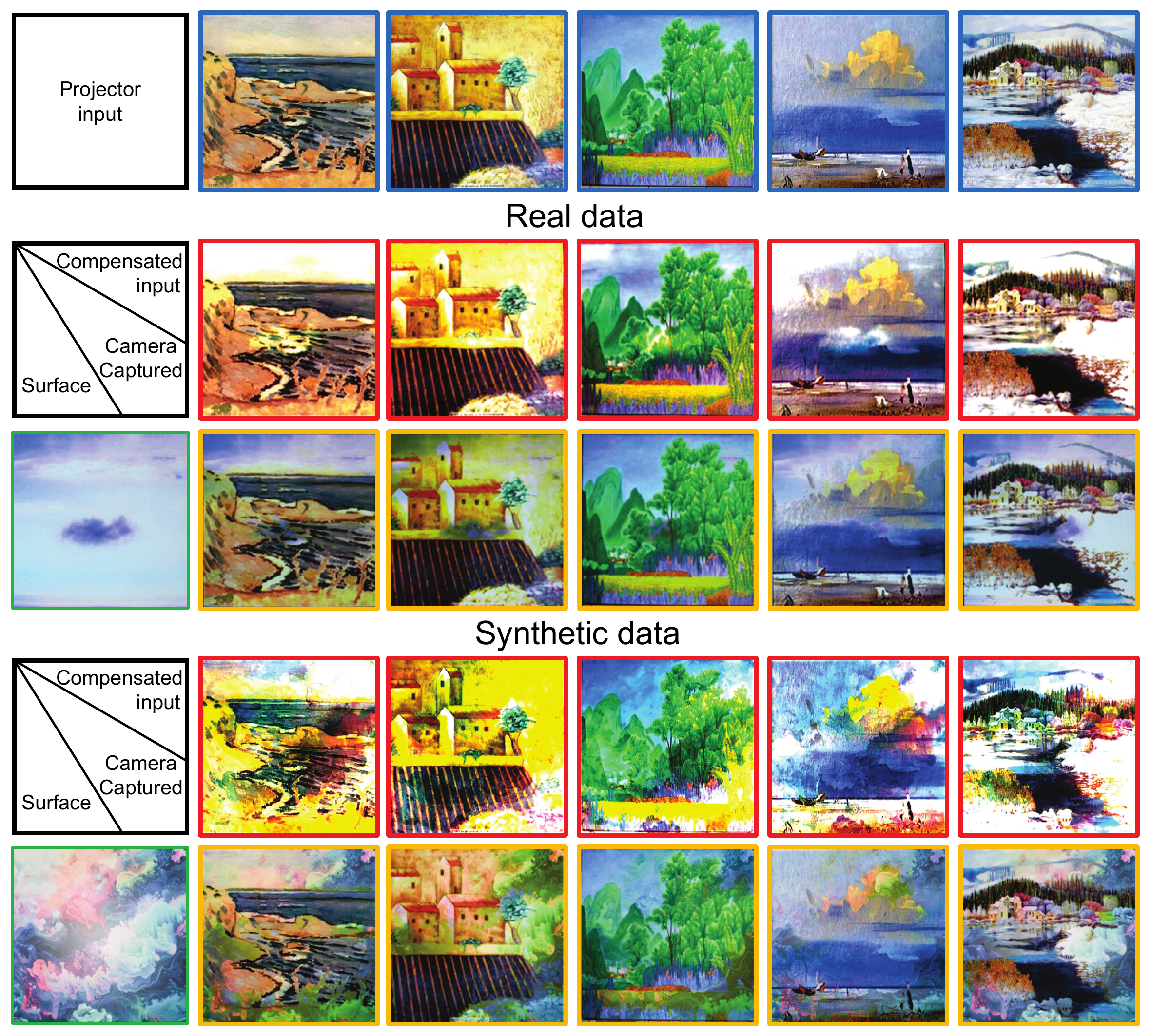}
\caption{Real and synthetic dataset. The 1\textsuperscript{st} row shows five different projector input images. The 2\textsuperscript{nd}  row is the compensated images of the 1\textsuperscript{st} row, given the surface in the 1\textsuperscript{st} column of the 3\textsuperscript{rd} row. The  2\textsuperscript{nd} to the 4\textsuperscript{th} columns of the 3\textsuperscript{rd} row are the corresponding camera-captured projection of the 2\textsuperscript{nd} row. Similarly, the last two rows are Blender synthesized images for another surface.}
\label{fig:data}
\end{figure}

To improve the practicability of PSA, we also built a synthetic dataset using Blender \cite{Blender2024} for model pre-training. 
Similarly to the real dataset, we also rendered 10 synthetic setups. For each setup, we provided 1,000 pairs of images for training and 200 for testing. Note that both the surface textures and the projector input differ from those in the real dataset. 
Some examples of images from our dataset are presented in \autoref{fig:data}. Our real and synthetic datasets can be leveraged in future work for model pre-training and network architecture exploration. 

\begin{table*}[htbp]
\centering
\caption{Quantitative evaluation of simulated captured compensation. Results are averaged over 10 real setups.}
\begin{tabular}{lccccccccc}
\toprule
 & \multicolumn{3}{c}{\textbf{\# Train = 8}} & \multicolumn{3}{c}{\textbf{\# Train = 48}} & \multicolumn{3}{c}{\textbf{\# Train = 125}}  \\ \cmidrule(lr){2-4} \cmidrule(lr){5-7} \cmidrule(lr){8-10}
 \textbf{Model}& \textbf{PSNR $\uparrow$} & \textbf{RMSE $\downarrow$} & \textbf{SSIM $\uparrow$} 
& \textbf{PSNR $\uparrow$} & \textbf{RMSE $\downarrow$} & \textbf{SSIM $\uparrow$} 
& \textbf{PSNR $\uparrow$} & \textbf{RMSE $\downarrow$} & \textbf{SSIM $\uparrow$} 
\\
\midrule
 PCNet & \textbf{26.7746} & \textbf{0.0805} & \textbf{0.8782} & \textbf{30.7357} & \textbf{0.0505} & \textbf{0.9268} & \textbf{31.1363} & \textbf{0.0483} & \textbf{0.9304} \\

 CompenNeSt reversed & 25.7832 & 0.0903 & 0.8327 & 29.9323 & 0.0554 & 0.9089 & 30.5871 & 0.0514 & 0.9161 \\
 \bottomrule
\end{tabular}
\label{tab:compensation_relighting}
\end{table*}

\begin{figure*}[!htbp]
\centering
\includegraphics[width=\linewidth]{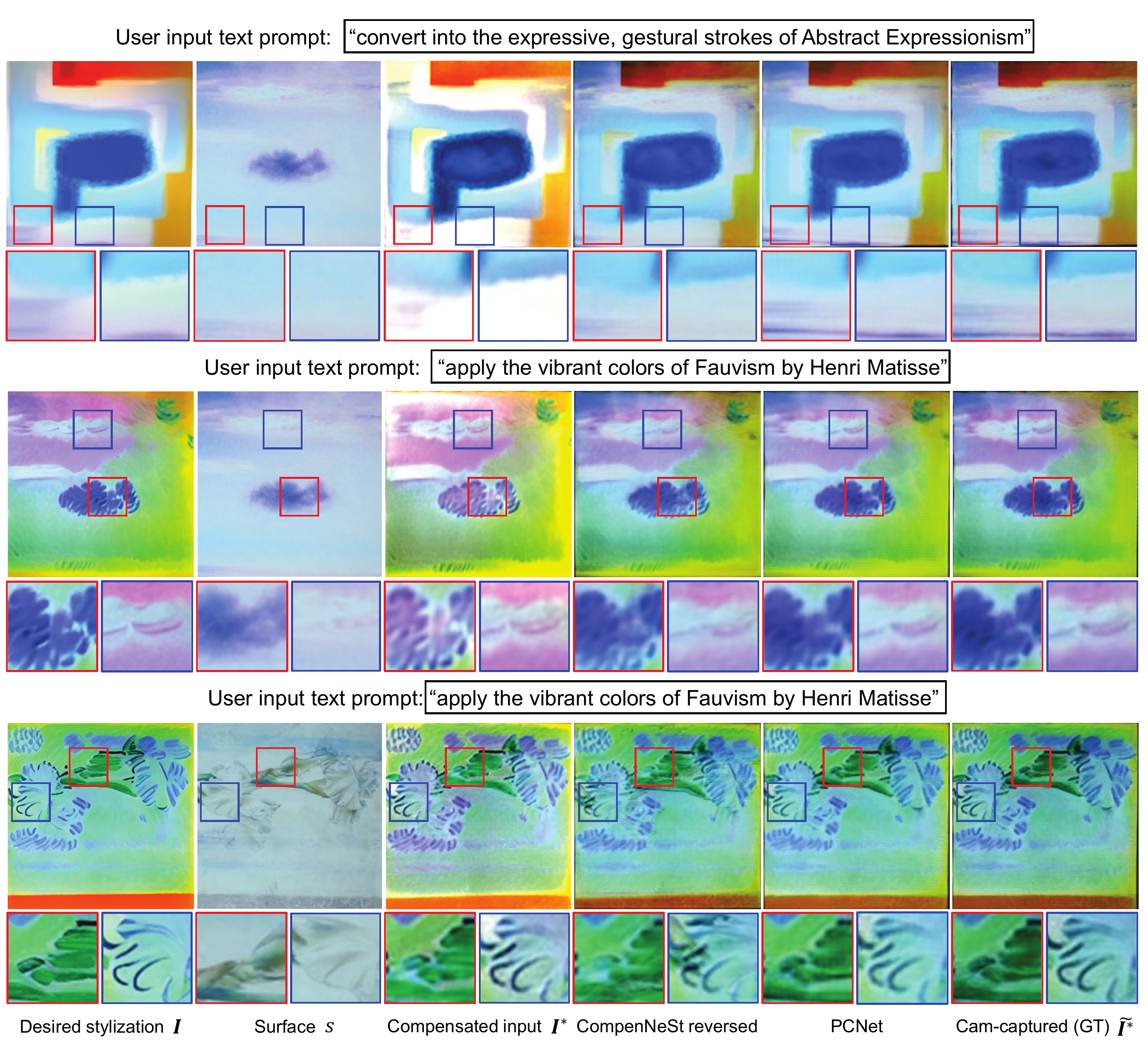}
\caption{Qualitative evaluation of simulated captured compensation. We provide three scenes in three rows, and each scene is under different projections generated by LGST. The first two rows share the same projection surface, but have different text prompts, while the last two rows share the same text prompt, but have different surface. 
The user input text prompts to stylize the surface are shown on the top of each row. 
The 1\textsuperscript{st} column is the desired viewer perceived effects, \ie, the 2\textsuperscript{nd} column stylized by LGST. The 2\textsuperscript{nd} column shows the original projection surfaces. The 3\textsuperscript{rd} column displays the compensated result of the 1\textsuperscript{st} column for projection. The 4\textsuperscript{th} and the 5\textsuperscript{th} columns are simulated captured projection by two methods. The last column presents real captured projection, \ie, the 3\textsuperscript{rd} column projected onto the 2\textsuperscript{nd} column. Comparing the 4\textsuperscript{th} and 5\textsuperscript{th} columns with the final column, PCNet demonstrates superior performance in simulating the project-and-capture process, surpassing CompenNeSt reversed in both color and texture accuracy.
Each image is provided with two zoomed-in patches for detailed comparison. More results are provided in \textbf{supplementary}.}
\label{fig:compensation_relighting}
\end{figure*}

\subsection{Simulated captured compensation}\label{sec:exp_simlulated_cmp}
Since the accuracy of the real project-and-capture simulation $\hat{\pi}$ and the compensation model $\pi^{\dagger}$ directly affects the final surface adaptation quality and the viewer’s perceived effects, we quantitatively and qualitatively evaluate the accuracy of the simulation by comparing the simulated camera-captured compensation, \ie, 
$\hat{\I^*}$ (\autoref{eq:simu-cap-cmp}) with the real camera-captured compensation, \ie, $\tilde{\I^*}$ (\autoref{eq:captured-compensation}). The results are shown in \autoref{tab:compensation_relighting} and \autoref{fig:compensation_relighting}. 

To explore different project-and-capture models, we adapted CompenNeSt \cite{huang2020CompenNeSt++} to simulate the project-and-capture process by swapping its input and output, and we name it \textbf{CompenNeSt reversed}. The advantage of our PCNet over this adapted CompenNeSt model is demonstrated by comparing the two methods in \autoref{tab:compensation_relighting} and \autoref{fig:compensation_relighting}. Clearly, our PCNet outperforms CompenNeSt reversed by a great margin in PSNR, RMSE, and SSIM, especially when the number of training images is smaller, \eg, \# Train = 8.

\begin{table*}[!h]
\centering
\setlength{\tabcolsep}{16pt} 
\caption{Quantitative comparison between language guided projector image generation (LAPIG) w/~ and w/o PSA. Each evaluation involved selecting 5 different textured projection surfaces in the 1\textsuperscript{st} column (images are shown in the bottom), then applying 10 same text prompts to guide surface stylization, and averaging the results from 50 trials.}
\label{tab:extended_psa}
\begin{tabular}{lcccccc}
\toprule
\multicolumn{1}{l}{} &
  \multicolumn{2}{c}{\textbf{PSNR $\uparrow$}} &
  \multicolumn{2}{c}{\textbf{RMSE $\downarrow$}} &
  \multicolumn{2}{c}{\textbf{SSIM $\uparrow$}} \\ \cmidrule(lr){2-3} \cmidrule(lr){4-5}  \cmidrule(lr){6-7} 
\textbf{Surface} &
  \multicolumn{1}{c}{w/~ PSA} &
  \multicolumn{1}{c}{w/o PSA} &
  \multicolumn{1}{c}{w/~ PSA} &
  \multicolumn{1}{c}{w/o PSA} &
  \multicolumn{1}{c}{w/~ PSA} &
  \multicolumn{1}{c}{w/o PSA} \\ \midrule
Bench  & \textbf{23.1286} & 20.6997 & \textbf{0.0728} & 0.0980 & \textbf{0.7013} & 0.6910 \\ 
Wood  & \textbf{22.2555} & 21.7816 & \textbf{0.0789} & 0.0860 & \textbf{0.6869} & 0.6599 \\ 
Stripe  & \textbf{20.1848} & 18.5286 & \textbf{0.1029} & 0.1246 & \textbf{0.7387} & 0.7316 \\ 
Sakura  & \textbf{15.8081} & 14.4061 & \textbf{0.1672} & 0.1964 & \textbf{0.3942} & 0.3772 \\ 
Spray  & \textbf{23.4177} & 22.0880 & \textbf{0.0708} & 0.0825 & \textbf{0.6654} & 0.6228 \\ 
\midrule
Average & \textbf{20.9589} & 19.5008 & \textbf{0.0985} & 0.1175 & \textbf{0.6373} & 0.6165 \\ 
\bottomrule
\end{tabular}
\includegraphics[width=0.87\linewidth]{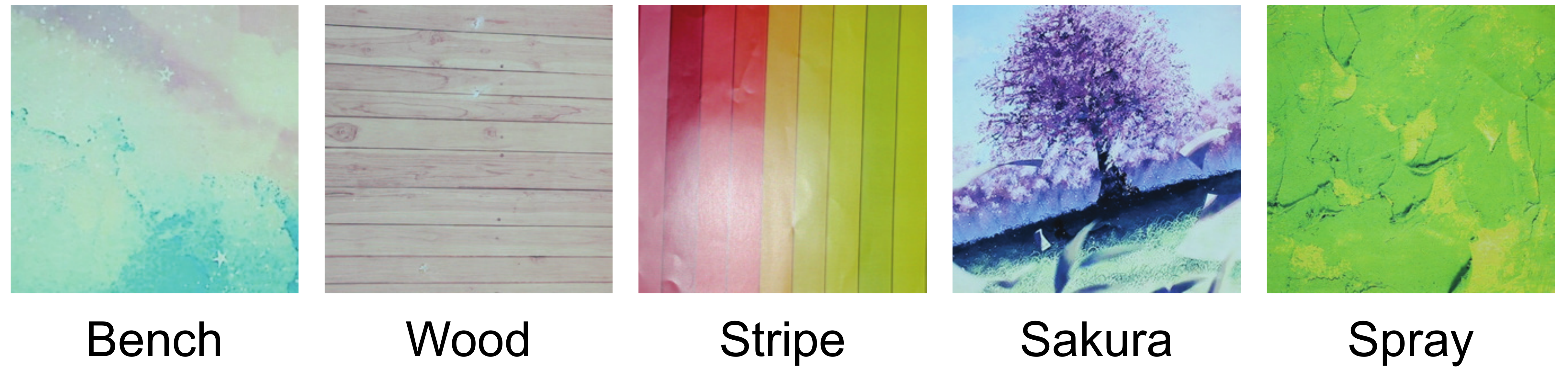}
\end{table*}

\subsection{Comparison of different PCNet training losses}\label{sec:exp_pcnet_losses}
The loss function for PCNet training is crucial for our LAPIG performance. Although pixel-wise $l_1$ and $l_2$ losses are commonly used to penalize pixel errors in various image reconstruction tasks, in \cite{huang2021deprocams}, Huang \etal~demonstrated that SSIM loss effectively recovers structural details in image compensation tasks. The performance of different combinations of the three loss functions is shown in \autoref{tab:relit_different_loss}. 

In summary, $l_1$ and $l_2$ losses alone produce suboptimal outcomes. Using $l_2$ loss alone may result in loss of detail, leading to lower visual quality compared to using $l_1$ loss. 
The combination of $l_1$, $l_2$, and SSIM losses yields the best performance during training, likely due to the balanced optimization it provides between fine-detail preservation, global structure alignment, and robustness to outliers. $l_1$ loss helps to retain pixel-wise fine details, $l_2$ loss suppresses large color errors, and SSIM loss improves structural details.

Given the potential biases and convergence issues associated with the use of individual loss functions \cite{zhao2017lossfunction}, we employed a combined loss of $l_1$ and $l_2$ during the early training stages to improve convergence, and changed to $l_1+l_2+\text{SSIM}$ to further improve fine details and structural details.

\begin{table}[!htb]
\centering
\caption{Comparison of different PCNet training losses. The results are averaged over 5 real setups and 5 synthetic setups.}
\begin{tabular}{lccc}
\toprule
\textbf{PCNet training loss} & \textbf{PSNR $\uparrow$} & \textbf{RMSE $\downarrow$} & \textbf{SSIM $\uparrow$} \\
\midrule
$l_1$ & 29.3504 & 0.0597 & 0.8783 \\
$l_2$ & 29.2355 & 0.0603 & 0.8848 \\
SSIM & 32.0635 & 0.0435 & 0.9485 \\
$l_1 + l_2$ & 32.4053 & 0.0418 & 0.9454 \\
$l_1 +$ SSIM & 33.2556 & 0.0379 & 0.9586 \\
$l_2 +$ SSIM & 33.7120 & 0.0360 & 0.9643 \\
$l_1 + l_2 + $ SSIM & \textbf{34.0953} & \textbf{0.0345} & \textbf{0.9662} \\
\bottomrule
\end{tabular}
\label{tab:relit_different_loss}
\end{table}

\begin{figure*}[!htbp]
\centering
\includegraphics[width=0.87\linewidth]{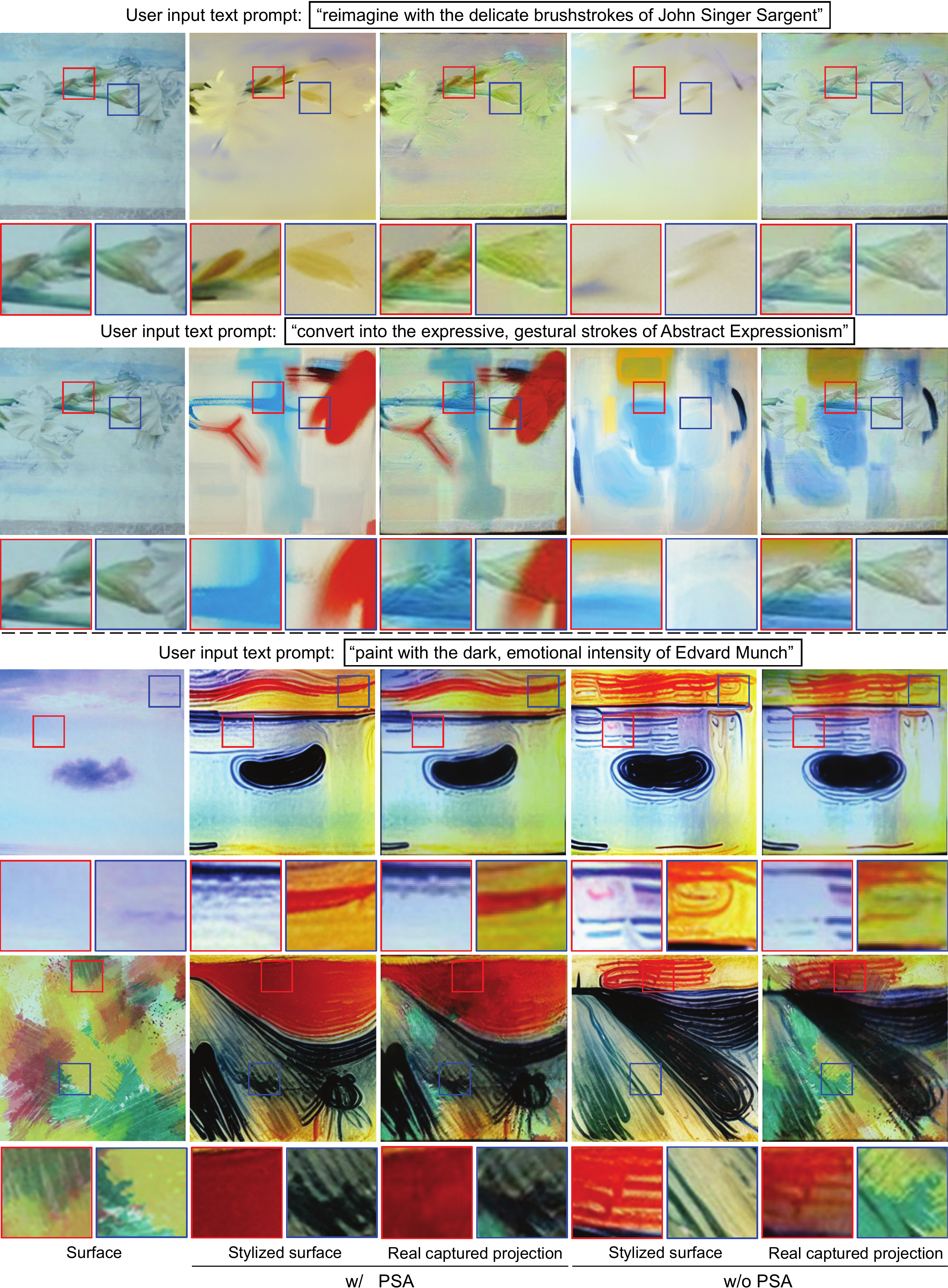}
\caption{Qualitative comparison between language guided projector image generation (LAPIG) w/ and w/o PSA. The first two rows show stylization of the \textit{same projection surface} using two \textit{different text prompts}, while the last two rows show stylization of \textit{two different projection surfaces} using the \textit{same text prompt}. 
The 1\textsuperscript{st} column shows the original projection surfaces. The 2\textsuperscript{nd} and 4\textsuperscript{th} columns are stylized projection surfaces by LAPIG w/ or w/o PSA, given the user input text prompt on the top of each surface(s). The 3\textsuperscript{rd} and 5\textsuperscript{th} columns present real captured scenes under projection w/ or w/o PSA, \ie, the 2\textsuperscript{nd} or 4\textsuperscript{th} column projected onto the 1\textsuperscript{st} column after projector compensation. Clearly, our PSA can adapt the stylization to the surface by showing fewer discrepancies between the target stylized surface (the 2\textsuperscript{nd} column) and the real captured projection (the 3\textsuperscript{rd} column). 
See \textbf{supplementary} for more results.}
\label{fig:cmp_psa}
\end{figure*}

\subsection{Effectiveness of PSA}

To show the effectiveness of our PSA, we compared the stylized surfaces generated w/ and w/o PSA, and their corresponding real captured projections. The quantitative and qualitative results are shown in \autoref{tab:extended_psa} and \autoref{fig:cmp_psa}, respectively. 
Clearly, PSA can adapt the stylization to the surface by showing fewer discrepancies between the target stylized surface (the 2\textsuperscript{nd} column) and the real captured projection (the 3\textsuperscript{rd} column), due to the projection consistency and color saturation losses (\autoref{eq:pc}  to \autoref{eq:ps}).
Comparing the 1\textsuperscript{st} and 2\textsuperscript{nd} columns with the 3\textsuperscript{rd} and 4\textsuperscript{th} columns
of \autoref{fig:cmp_psa}, it is obvious that our LAPIG (w/ PSA) tends to generate image content that fits/adapts the projection surfaces, \eg, generates dark stylization around the hard-to-compensate dark surface regions, while without PSA, the LGST generated stylization tends to ignore these hard-to-compensate regions, resulting in artifacts.

The effectiveness of PSA is also demonstrated in \autoref{tab:different_psa_loss}, where w/ PSA shows superior PSNR, RMSE and SSIM compared to w/o PSA on 3 different surfaces with 5 different text prompts.

\subsection{Effectiveness of different PSA losses}

We evaluated the effectiveness of different PSA losses in \autoref{eq:pc} to \autoref{eq:ps}. Specifically, we progressively removed or used a combination of projection consistency loss $\mathcal{L}_\text{pc}$, projection saturation loss $\mathcal{L}_\text{ps}$, and compensation saturation loss $\mathcal{L}_\text{cs}$. The quality of the stylized surfaces was evaluated by PSNR, RMSE, SSIM, and the average number of optimization iterations required for convergence (Mean iter.). The results are shown in \autoref{tab:different_psa_loss}, compared with w/o PSA, it is clear that projection saturation loss, projection consistency loss and compensation saturation loss can improve the quality of the stylized image. In particular, the compensation saturation loss and projection saturation loss can  significantly improve convergence, \eg, Mean iter. drops drastically from 86 to around 20. Moreover, combining all three losses achieves a balance between  surface stylization quality and convergence. Note that besides LAPIG, the three losses can also be applied to general projection mapping applications, such as ProCams simulation and projector compensation, to improve both quality and convergence.

\begin{table}[!htbp]
\centering
\caption{Comparisons of different PSA losses. Each loss evaluation involved selecting 3 different textured projection surfaces, then performing 5 same text prompts guided surface stylization, and averaging the results from 15 trials.}
\resizebox{\linewidth}{!}{
\begin{tabular}{lccccc}
\toprule
\textbf{PSA loss} & \textbf{PSNR $\uparrow$} & \textbf{RMSE $\downarrow$} & \textbf{SSIM $\uparrow$} & \textbf{Mean iter.} \\
\midrule
$\mathcal{L}_\text{pc}$ & 19.9299 & 0.1038 & 0.7447 & 86 \\ 
$\mathcal{L}_\text{cs}$ & \textbf{20.6526} & 0.0948 & 0.7433 & 19 \\ 
$\mathcal{L}_\text{ps}$ & 20.3378 & 0.0978 & 0.7487 & \textbf{17} \\ 
$\mathcal{L}_\text{ps}+\mathcal{L}_\text{cs}$ & 20.5002 & 0.0958 & 0.7488 & \textbf{17} \\ 
$\mathcal{L}_\text{pc}+\mathcal{L}_\text{ps}+\mathcal{L}_\text{cs}$ & 20.5831 & \textbf{0.0948} & \textbf{0.7509} & 21 \\ 
\midrule
w/o PSA & 18.8891 & 0.1158 & 0.7149 & 0 \\ 
\bottomrule
\end{tabular}
}
\label{tab:different_psa_loss}
\end{table}

\section{Discussion}\label{sec:discussion}

\subsection{Applicability to other settings}

For museums and galleries, LAPIG can be used to design the surfaces of exhibited objects in an artistic way that enhances the user experience. For LGST, we use InstructPix2Pix \cite{brooks2023instructpix2pix}, which is based on the stable diffusion model \cite{rombach2022high}. In theory, other text-guided image-to-image models are also applicable to our LAPIG.

\begin{figure}[!htbp]
\centering
\includegraphics[width=\linewidth]{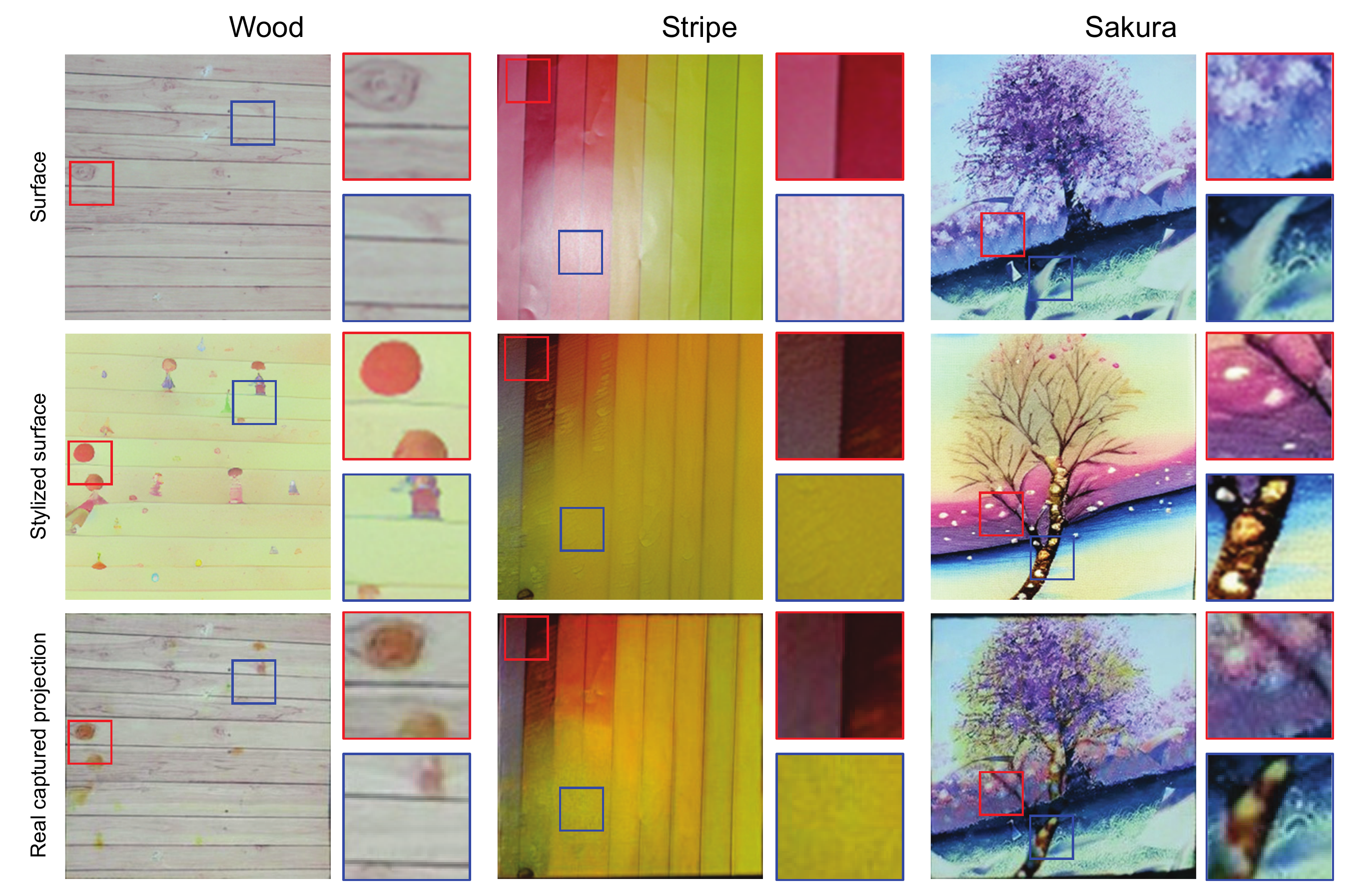}
\caption{Failure cases of our LAPIG. Three user input text prompts are used to stylize three projection surfaces: \textbf{Render it as a whimsical children's book illustration} for the \textit{Wood} surface, and \textbf{Render it in the style of a Van Gogh oil painting} for the \textit{Stripe} surface, and \textbf{Imagine it as a traditional Indian miniature painting} for the \textit{Sakura} surface, respectively.  
The 1\textsuperscript{st} row shows the original projection surfaces. The 2\textsuperscript{nd} row is stylized projection surfaces by LAPIG. The 3\textsuperscript{rd} row presents real captured scenes under projection, \ie, the 2\textsuperscript{nd} row projected onto the 1\textsuperscript{st} row after projector compensation. 
Note that for \textit{Wood} and \textit{Sakura}, the real captured projections in the 3\textsuperscript{rd} look different from the 2\textsuperscript{nd} row, indicating failed projection surface adaptation. For \textit{Stripe}, although the 2\textsuperscript{nd} and 3\textsuperscript{rd} rows look similar, but the LGST generated stylization failed to represent ``Van Gogh oil painting style".}
\label{fig:extended_psa_3}
\end{figure}

\subsection{Limitations and future work}
\vspace{1mm}\noindent\textbf{Failure cases of our LAPIG}  are shown in \autoref{fig:extended_psa_3}. To address these issues, future work could focus on improving the accuracy of project-and-capture simulation and projector compensation techniques, along with designing and training LGST specifically for projection mapping applications.

\vspace{1mm}\noindent\textbf{Low resolution.} To achieve fast performance in our LAPIG tasks, we currently process images at low resolutions. Although our LAPIG  can be directly applied to projectors and cameras with higher resolutions without modifying the neural network architecture, this could lead to increased processing times and GPU memory. Therefore, efficient network architectures, \eg, CompenHR \cite{wang2023CompenHR} may be applied to address this problem.

\vspace{1mm}\noindent\textbf{Basic language guidance.} The current LAPIG framework is limited to basic text interaction and applying a global style transfer to the entire projection surface. Occasionally, it also minimally and locally adapts the stylization to the surface with slight transformations (\autoref{fig:extended_psa_3}). In addition, real-world use cases frequently include complex user interactions that integrate visual, auditory, and tactile modalities. Expanding LAPIG to accommodate multimodal interactions will align it more closely with the user's intention, resulting in more intuitive and engaging experiences. 

\vspace{1mm}\noindent\textbf{User study.}
Our LAPIG lacks a direct assessment of viewer perceptual satisfaction. Future work will incorporate thorough user studies to assess and refine LAPIG. 

\vspace{1mm}\noindent\textbf{Complex scenes.} 
Our LAPIG was not tested in a wide variety of scenarios, and the current approach focuses mainly on modifying projection surface textures. However, real-world conditions typically involve more complexity, such as different depths, geometries, and lighting conditions. In future work, we plan to expand our method to tackle more complex scenes, where the difficult shape and indirect light conditions are involved. Extending LAPIG to dynamic projection mapping is also an interesting direction to explore.

\section{Conclusion}\label{sec:conclusion}

In this paper, we introduced LAPIG, a novel language guided projector image generation approach for surface adaptation and stylization. Our LAPIG framework can generate a projector image that follows the user's instruction while reducing projection surface artifacts, providing support for extended applications in various fields such as artistic and animated projection mapping. We also introduce projection consistency and color saturation losses to guide projection surface adaptation (PSA), and they are expected to facilitate other projection mapping applications. 

\acknowledgments{
We thank the anonymous reviewers for valuable and inspiring comments and suggestions. 
}

\bibliographystyle{abbrv-doi-hyperref}

\bibliography{LAPIG_arxiv}

\clearpage

\maketitlesupplementary
\appendix  
\setcounter{page}{1}

\vspace*{-1.2cm}

\section{Introduction}

\begin{figure}[!htbp]
\centering
\includegraphics[width=\linewidth]{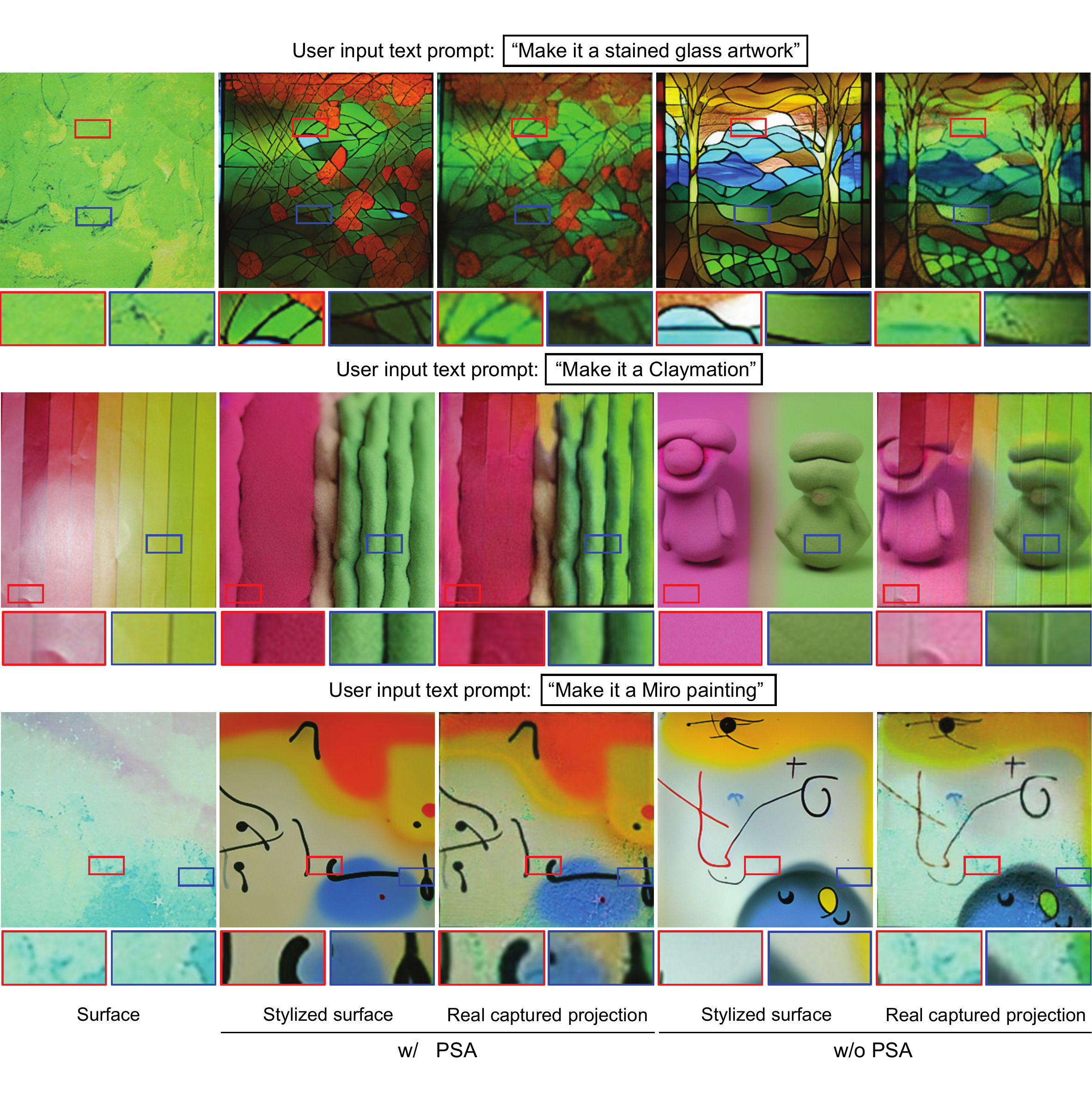}
\vspace*{-0.8cm}
\caption{Comparison between language guided projector image generation (LAPIG) w/ and w/o PSA. \textbf{Three} user input text prompts are used to stylize \textbf{three} projection surfaces. The text prompt to stylize the surfaces shown here are \textbf{Make it a stained glass artwork} for the 1\textsuperscript{st} row, and \textbf{Make it a Claymation} for the 2\textsuperscript{nd} row, and \textbf{Make it a Miro painting} for the 3\textsuperscript{rd} row. The 1\textsuperscript{st} column shows the original projection surfaces. The 2\textsuperscript{nd} and 4\textsuperscript{th} columns are stylized projection surfaces by LGST w/ or w/o PSA, given the user input text prompt on the top of each surface. The 3\textsuperscript{rd} and 5\textsuperscript{th} columns present real captured scene under projection w/ or w/o PSA, i.e., the 2\textsuperscript{nd} or 4\textsuperscript{th} column projected onto the 1\textsuperscript{st} column after projector compensation.}
\label{fig:extended_psa_1}
\end{figure}

\begin{figure}[!htbp]
\centering
\includegraphics[width=\linewidth]{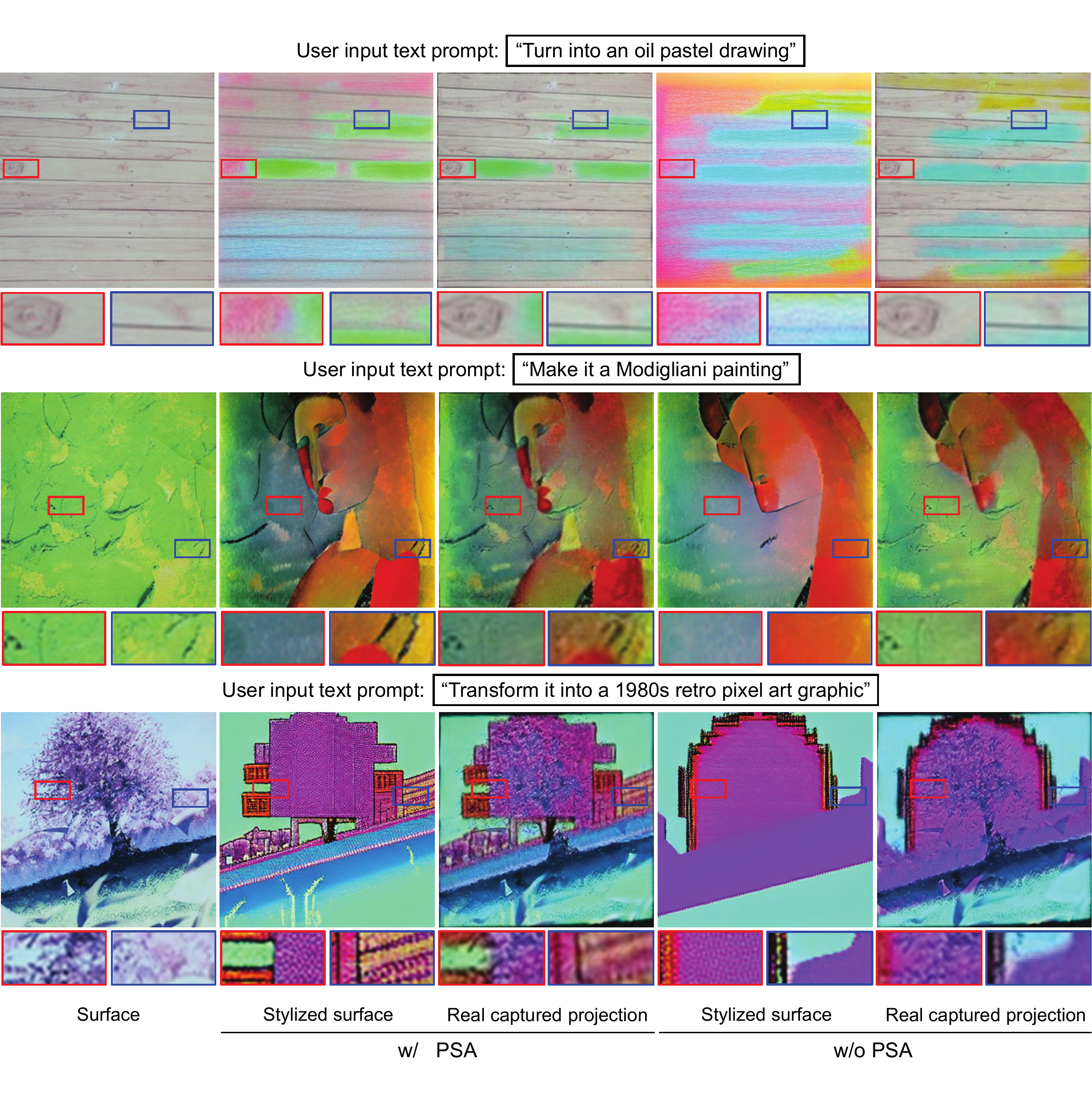}
\caption{Comparison between language guided projector image generation (LAPIG) w/ and w/o PSA. \textbf{Three} user input text prompts are used to stylize \textbf{three} projection surfaces. The text prompt to stylize surfaces shown here are \textbf{Turn into an oil pastel drawing} for the 1\textsuperscript{st} row, and \textbf{Make it a Modigliani painting} for the 2\textsuperscript{nd} row, and \textbf{Transform it into a 1980s retro pixel art graphic} for the 3\textsuperscript{rd} row. The 1\textsuperscript{st} column shows the original projection surfaces. The 2\textsuperscript{nd} and 4\textsuperscript{th} columns are stylized projection surfaces by LGST w/ or w/o PSA, given the user input text prompt on the top of each surface. The 3\textsuperscript{rd} and 5\textsuperscript{th} columns present real captured scene under projection w/ or w/o PSA, i.e., the 2\textsuperscript{nd} or 4\textsuperscript{th} column projected onto the 1\textsuperscript{st} column after projector compensation.}
\label{fig:extended_psa_2}
\end{figure}

In this supplementary material, we show additional experimental results, including more surfaces and text prompts in \autoref{fig:extended_psa_1} and \autoref{fig:extended_psa_2}. More qualitative comparisons between language guided projector image generation w/ and w/o PSA are shown in \autoref{fig:extended_psa_1} and \autoref{fig:extended_psa_2}. The results further demonstrate the effectiveness of our PSA, because w/ PSA outperforms w/o PSA by a significant margin. 

\vspace*{-0.3cm}

\vspace*{-1.6cm}

\end{document}